\documentclass[10pt,twocolumn,letterpaper]{article}
\usepackage[pagenumbers]{wacv}

\usepackage{graphicx}
\usepackage{amsmath}
\usepackage{amssymb}
\usepackage{booktabs}
\usepackage{multirow}
\usepackage{subfiles}

\usepackage[pagebackref,breaklinks,colorlinks]{hyperref}
\usepackage[accsupp]{axessibility} 

\usepackage[capitalize]{cleveref}
\crefname{section}{Sec.}{Secs.}
\Crefname{section}{Section}{Sections}
\Crefname{table}{Table}{Tables}
\crefname{table}{Tab.}{Tabs.}

\begin{document}

\title{ORFormer: Occlusion-Robust Transformer for Accurate Facial Landmark Detection}
\author{Jui-Che Chiang$^{1,3}$ \quad Hou-Ning Hu$^{2}$ \quad $^*$Bo-Syuan Hou$^{1}$ \quad $^*$Chia-Yu Tseng$^{1}$ \\ \quad Yu-Lun Liu$^1$ \quad Min-Hung Chen$^{3}$ \quad Yen-Yu Lin$^{1}$ \vspace{0.3em} \\
{\normalsize $^1$National Yang Ming Chiao Tung University} \quad
{\normalsize $^2$MediaTek Inc.} \quad 
{\normalsize $^3$NVIDIA}\\
{\url{https://ben0919.github.io/ORFormer}}
}
\maketitle
\def\thefootnote{*}\footnotetext{means equal contribution}

\maketitle

\newcommand{\jc}[1]{\textcolor{red}{#1}}

\def\ourmethod{ORFormer}
\def\ourtoken{messenger token}
\def\ourembed{messenger embedding}
\def\Ourtoken{Messenger token}
\def\Ourembed{Messenger embedding}
\begin{abstract}
Although facial landmark detection (FLD) has gained significant progress, existing FLD methods still suffer from performance drops on partially non-visible faces, such as faces with occlusions or under extreme lighting conditions or poses. 
To address this issue, we introduce \ourmethod, a novel transformer-based method that can detect non-visible regions and recover their missing features from visible parts. 
Specifically, \ourmethod~associates each image patch token with one additional learnable token called the \ourtoken.
The \ourtoken~aggregates features from all but its patch.
This way, the consensus between a patch and other patches can be assessed by referring to the similarity between its regular and \ourembed s,  enabling non-visible region identification.
Our method then recovers occluded patches with features aggregated by the messenger tokens.
Leveraging the recovered features, \ourmethod~compiles high-quality heatmaps for the downstream FLD task. 
Extensive experiments show that our method generates heatmaps resilient to partial occlusions. 
By integrating the resultant heatmaps into existing FLD methods, our method performs favorably against the state of the arts on challenging datasets such as WFLW and COFW.
\end{abstract}

\section{Introduction} 
Facial landmark detection (FLD) aims to localize specific key points on human faces, such as those on eyes, noses, and mouths. 
It is pivotal for numerous downstream applications, such as face recognition \cite{parkhi2015deep, juhong2017face}, facial expression recognition \cite{barsoum2016training, mollahosseini2016going}, head pose estimation \cite{hempel20226d, valle2020multi}, and augmented reality \cite{kang2021real, wei2021assessing}.
Recent advances in deep neural networks have significantly enhanced facial landmark detection \cite{feng2018wing,wang2019adaptive,guo2019pfld,huang2021adnet,xia2022sparse,zhou2023star}.
However, existing FLD methods suffer from performance drops on partially non-visible faces caused by occlusions, extreme lighting conditions, or extreme head rotations, because the features extracted from non-visible regions are corrupted.  
An FLD method with non-visible region detection and reliable feature extraction is in demand.

\begin{figure*}[t!]
    \centering
    \includegraphics[width=1\textwidth]{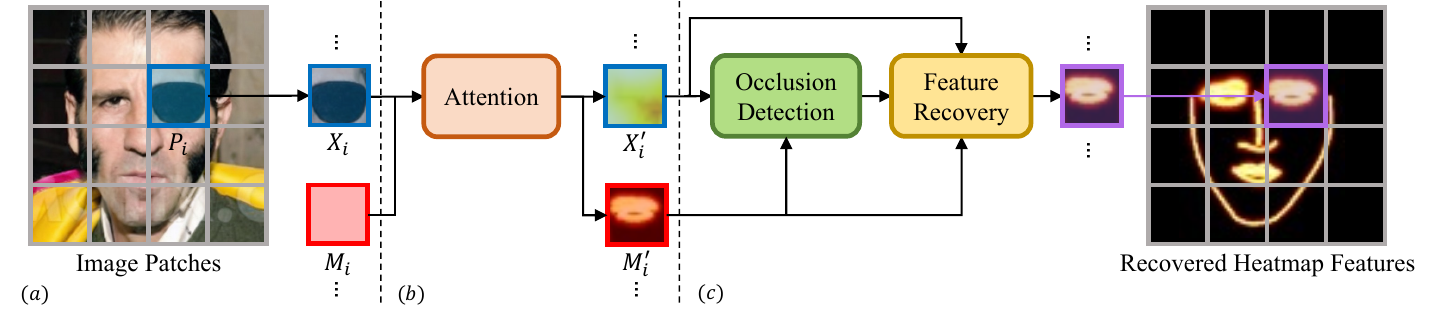}
    \caption{
    \textbf{Overview of our \ourmethod.}
    (a) For each patch $P_i$, we introduce a patch token $X_i$ and a learnable \ourtoken~$M_i$ for occlusion detection and handling.
    (b) The \ourtoken~computes attention with patch tokens other than its corresponding one. 
    (c) We detect occlusion by evaluating the dissimilarity between the regular embedding $X_i'$ and the \ourembed~$M_i'$, and then recover occluded features based on the \ourembed~which is aggregated from other image patches, if occlusion is present in patch $P_i$.
    }
    \label{fig:teaser}
\end{figure*}

In this work, we introduce an occlusion-robust transformer, called {\em ORFormer}, which can identify non-visible regions and recover their missing features, and is applied to generate high-fidelity heatmaps resilient to challenging scenarios.
As illustrated in Figure~\ref{fig:teaser}, our ORFormer builds on Vision Transformer \cite{dosovitskiy2020image}, where image patch tokens interact with each other via the self-attention mechanism.
For non-visible part detection, we associate each patch token $X_i$ with an extra learnable token $M_i$ called {\em \ourtoken}.

The \ourtoken~$M_i$ simulates occlusion present in patch $i$ and aggregates features from all patch tokens except $X_i$.
Subsequently, our occlusion detection module accesses the disparity between the regular patch embedding $X_i'$ and the \ourembed~$M_i'$ to determine if occlusion is present in patch $i$.
For occlusion handling, our feature recovery module recovers the missing features of the occluded patch by a convex combination of $X_i'$ and $M_i'$ with the combination coefficient predicted by the occlusion detection module. 
The resulting features are then utilized to generate heatmaps, and our proposed mechanism makes the output heatmaps remain robust in extreme scenarios.

We integrate the high-quality heatmaps generated by ORFormer as complementary information into existing landmark detection methods~\cite{huang2021adnet,zhou2023star}. 
Our method achieves state-of-the-art performance on multiple benchmark datasets, including WFLW~\cite{wu2018look} and COFW~\cite{burgos2013robust}, showcasing the robustness of our method in handling partially non-visible faces.

The main contributions of this work are summarized as follows:
First, we present a novel occlusion-robust transformer, ORFormer, which utilizes the proposed learnable \ourtoken~to simulate potential occlusions and recover missing features.
ORFormer enables a transformer to detect and handle non-visible tokens in a general way.
Second, our ORFormer is applied for robust heatmap generation, thereby enhancing the applicability of existing FLD methods to partially non-visible faces.
Third, our method performs favorably against state-of-the-art facial landmark detection methods on two challenging datasets, WFLW and COFW, showcasing its robustness to extreme cases
\section{Related Work}
\subsection{Facial Landmark Detection}
Most FLD methods rely on coordinate regression and/or heatmap regression. 
The former directly estimates the landmark coordinates of a face. 
The latter predicts a heatmap for each landmark and completes FLD with post-processing.

\paragraph{Coordinate Regression.}
Some methods \cite{feng2018wing,guo2019pfld,wu2018look,zhang2014facial, zhou2013extensive, miao2018direct} employ linear layers as decoders to regress landmarks from CNN features. 
Feng~\etal~\cite{feng2018wing} design a new loss function for improved landmark supervision.
Wu~\etal~\cite{wu2018look} utilize facial contours to impose constraints on landmark supervision while providing a dataset with various extreme cases.
Miao~\etal~\cite{miao2018direct} proposed Fourier feature pooling to handle highly nonlinear relationships between images and facial shapes.
These methods offer end-to-end trainable solutions.

To leverage the self-attention mechanism in Transformer~\cite{vaswani2017attention} for facial structures exploration, some studies~\cite{li2022towards,li2022repformer,watchareeruetai2022lotr,li2024cascaded, wu2022transmarker,xia2022sparse} utilize the transformer decoder to learn the mapping between CNN features and landmarks. 
Xia~\etal~\cite{xia2022sparse} propose a coarse-to-fine decoder focusing on sparse local patches. 
Li~\etal~\cite{li2022repformer} learn landmark queries along pyramid CNN features. 
However, linear layers in CNN and global feature dependence in transformers are sensitive to partial occlusions.

\paragraph{Heatmap Regression.}
Inspired by the advances in heatmap generation~\cite{ronneberger2015u, newell2016stacked, sun2019deep}, some studies~\cite{bulat2021subpixel, dong2018supervision,kumar2020luvli, newell2016stacked,nibali2018numerical,wang2019adaptive} integrate heatmap regression into facial landmark detection.
They convert landmark annotations to heatmaps for model supervision. 
Kumar~\etal~\cite{kumar2020luvli} estimate uncertainty and visibility likelihood with heatmaps for stable model convergence. 
Newell~\etal~\cite{newell2016stacked} employ a stacked hourglass network with intermediate heatmap supervision and utilize {\tt Argmax} operator to obtain landmarks. 
However, Argmax in heatmap regression limits direct supervision by landmarks due to its non-differentiable nature.

Recent studies, such as replacing Argmax with other differential decoders, enable heatmap regression methods to be end-to-end trainable and supervised by both heatmaps and landmarks.
For example, Jin~\etal~\cite{jin2021pixel} reduce heatmap regression to confidence score and offset prediction to avoid heavy upsampling layers and the use of Argmax.
With the aid of differential decoders, Huang~\etal~\cite{huang2021adnet} and Zhou~\etal~\cite{zhou2023star} design new loss functions with both landmark and heatmap supervision to alleviate the negative impact caused by landmark annotation ambiguities. 
Micaelli~\etal~\cite{micaelli2023recurrence} utilize the deep equilibrium model to compute cascaded landmark refinement. 
The capability of heatmap regression methods that can be supervised by both landmarks and heatmaps while preserving facial structures has propelled them to the state-of-the-art status.

However, the aforementioned coordinate regression and heatmap regression methods are vulnerable to faces with partial occlusions, under extreme lighting conditions, or in extreme head rotations due to feature occlusion and corruption. 
 \begin{figure*}[t!]
    \centering
    \includegraphics[width=0.98\textwidth]{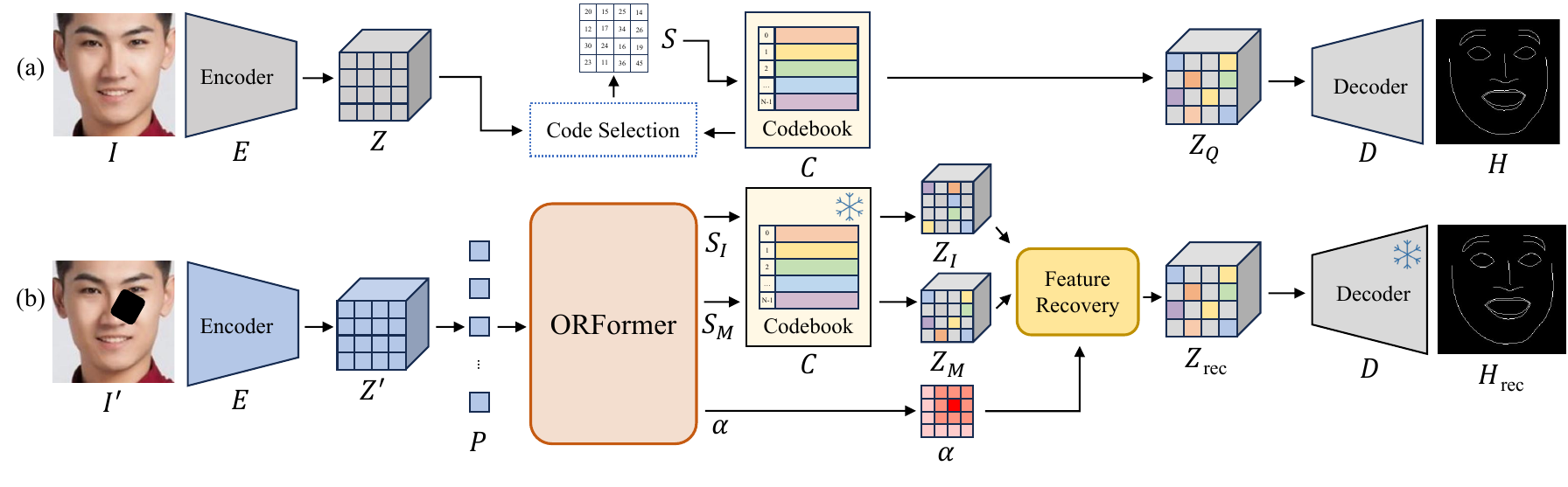}
    \caption{\textbf{Overview of our method.} (a) We first train a quantized heatmap generator, which takes an image $I$ as input and generates its edge heatmaps $H$.
    After pre-training, the prior knowledge of unoccluded faces is encoded in the codebook $C$ and decoder $D$. 
    (b) With the frozen codebook and decoder, we introduce \ourmethod~to generate the occlusion map $\alpha$ and two code sequences $S_I$ and $S_M$, leading to quantized features $Z_I$ and $Z_M$. 
    The recovered feature $Z_\text{rec}$ is yielded by merging $Z_I$ and $Z_M$ with patch-specific weights given in $\alpha$, and is used to produce occlusion-robust heatmaps $H_\text{rec}$.
    }
    \label{fig:overview}
\end{figure*}

\subsection{Occlusion-Robust Facial Landmark Detection}

We discuss three major categories of methodologies for occlusion-robust facial landmark detection as follows:

Methods in the first category such as \cite{kumar2020luvli,li2024cascaded,newell2016stacked,wu2015robust} estimate the probability of occlusion occurrence for each landmark and alleviate the negative impact of corrupted features computed in occluded areas.
For example, Kumar \etal \cite{kumar2020luvli} and Li \etal \cite{li2024cascaded} propose joint landmark location, uncertainty, occlusion probabilities, and/or visibility prediction. 
However, these methods rely on additional annotations indicating whether a landmark is occluded or not, while our method does not require these annotations.

The second category explores the consensus among image patches to identify occluded ones.
For example, Burgos-Artizzu \etal \cite{burgos2013robust} propose a method that enforces regressors focusing on different image patches to reach a consensus, trusting those using local features from non-occluded areas.
While their method and ours share a similar concept, their method ignores the occluded features without recovering them, restricting the ability of occluded landmark detection.
In contrast, we propose the \ourtoken, which aggregates information from non-occluded areas and enables feature recovery for occluded patches.

The third category utilizes global context to deal with occlusions.
Merget \etal \cite{merget2018robust} introduce global context directly into a fully convolutional neural network.
Zhu \etal \cite{zhu2019robust} propose a geometry-aware module to excavate geometric relationships between different facial components, while Zhu \etal \cite{zhu2022occlusion} model the hierarchies between facial components. 
However, these works do not explicitly detect the occluded areas and therefore do not recover features for these areas, being suboptimal for significant occlusions.

\subsection{Transformer for Feature Recovery}
Transformers~\cite{dosovitskiy2020image,vaswani2017attention} have been widely adopted in vision tasks.
Transformers leverage attention mechanisms to capture long-range dependencies between tokens, but are sensitive to feature corruption or partial occlusions.

To address this issue, Xu~\etal~\cite{xu2022learning} utilize cross-attention to recover occluded features between different frames in the context of object re-identification. 
However, their method relies on multiple frames, whereas our approach focuses on recovering occluded features within a single image.
Park~\etal~\cite{park2022handoccnet} proposes a method for 3D hand mesh estimation that involves training a CNN block to separate primary and secondary features, followed by utilizing cross-attention to recover occluded features. 
In contrast, our method integrates occlusion detection and handling mechanisms into a single transformer, enabling adaptive detection and recovery of occluded features within a single frame.

Zhou~\etal~\cite{zhou2022towards} pre-train a quantized autoencoder \cite{esser2021taming}, employ a ViT model \cite{dosovitskiy2020image}, and utilize self-attention to recover corrupted features for blind face restoration.
While their approach shares similarities with ours, relying on self-attention with partially corrupted features may fail since attention values of the occluded tokens cannot be faithfully computed. 
To alleviate this issue, we develop messenger tokens and present a module to adaptively combine the regular and messenger embeddings for feature recovery.

\section{Proposed Method}
The section presents \ourmethod, a general method that can be integrated into a regular transformer for occlusion detection and handling.
Figure~\ref{fig:overview} illustrates our method.
Firstly, we adopt the concept of vector quantization \cite{van2017neural}, similar to the approach in Codeformer \cite{zhou2022towards}, and pre-train a quantized heatmap generator (Section~\ref{section:VQVAE}). 
Subsequently, the learned discrete codebook and decoder are employed as a prior for heatmap generation. 
Leveraging this learned prior, we utilize \ourmethod~for code sequence prediction and feature recovery for the partially occluded image patches (Section~\ref{section:\ourmethod}).
Lastly, with the aid of \ourmethod, we integrate our heatmaps generated from recovered features into the existing FLD methods (Section~\ref{section:FLD method}).

\begin{figure*}[t!]
    \centering
    \includegraphics[width=0.85\textwidth]{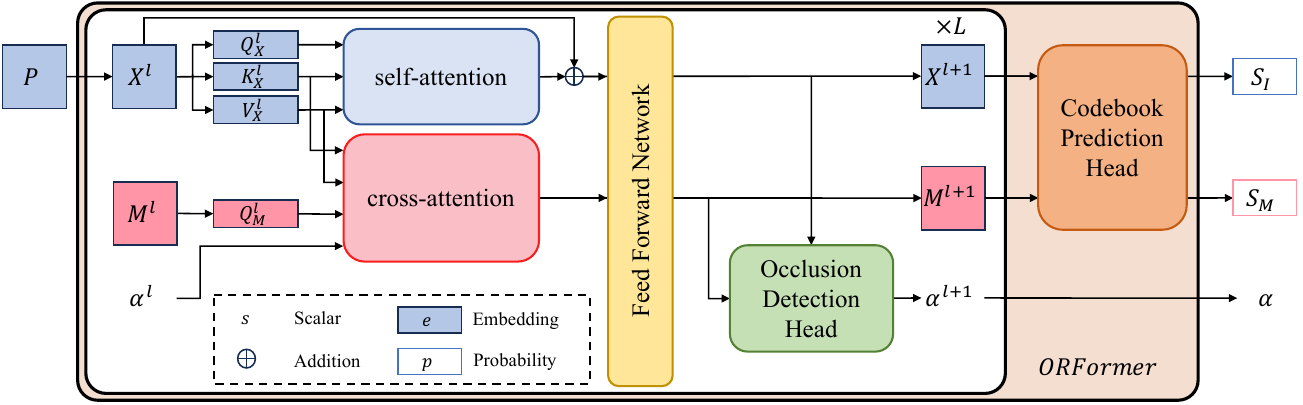}
    \caption{
    \textbf{Network architecture of \ourmethod.} 
    \ourmethod~takes image patches $P$ as input and generates two code sequences $S_I$ and $S_M$ via the codebook prediction head. 
    While $S_I$ is computed by referring to the image patch tokens, $S_M$ is by the messenger tokens.
    The occlusion map $\alpha$ represents the patch-specific occlusion likelihood and is inferred by the occlusion detection head.
    }
    \label{fig:method}
\end{figure*}
\subsection{Quantized Heatmap Generator}\label{section:VQVAE}

To enhance robustness against occlusions during heatmap generation, we include the training of a quantized heatmap generator. 
By training this generator on faces without occlusions, we can learn a high-dimensional latent space tailored explicitly for heatmap generation under ideal conditions. 
With the learned codebook, we reduce uncertainty in restoring occluded features, as the code items are learned from non-occluded faces.

As illustrated in Figure~\ref{fig:overview}(a), an unoccluded face image $I \in \mathbb{R}^{h \times w \times 3}$ is encoded into the latent space $Z \in \mathbb{R}^{m \times n \times d}$ by an encoder $E$. 
Following the principles in VQVAE \cite{van2017neural}, each patch $Z^{i,j}$ in the encoded features $Z$ is replaced with the nearest dictionary item, \ie, code, in the learnable codebook $C=\{c_s \in \mathbb{R}^d\}_{s=0}^{N-1}$ of $N$ codes to obtain the quantized feature $Z_Q \in \mathbb{R}^{m \times n \times d}$ and its corresponding code index sequence $S \in \{0, 1, ..., N-1\}^{h \times w}$, \ie,
\begin{equation}\label{equation:code replacement}
\begin{aligned}
    Z_Q^{i, j} &= \arg\min_{c_s \in C} ||Z^{i,j} - c_s||_2\mbox{~~and~~}\\
    S^{i,j} &= \arg\min_{s} ||Z^{i,j} - c_s||_2.
\end{aligned}
\end{equation}

Subsequently, the decoder $D$ generates the edge heatmaps $H \in \mathbb{R}^{h \times w \times N_E}$ based on the quantized feature $Z_Q$, where $N_E$ is the number of edges (facial contours) per face. 
In this work, we adopt the same edge heatmap definition as that in \cite{wu2018look}.

\paragraph{Loss Functions.}
To train the quantized heatmap generator with a learnable codebook, we utilize an image-level loss $\mathcal{L}_\text{img}$. 
In addition, we incorporate an intermediate latent space loss $\mathcal{L}_\text{latent}$ to minimize the distance between the codebook $C$ and the encoded feature $Z$.
These loss functions are defined by
\begin{equation}\label{equation:codebook loss}
\begin{aligned}
    \mathcal{L}_\text{img} &= ||H - \hat{H}||_2^2\mbox{~~and~~}\\
    \mathcal{L}_\text{latent} &= ||{\tt sg}(Z) - Z_Q||_2^2 + \beta||Z - {\tt sg}(Z_Q)||_2^2,
\end{aligned}
\end{equation}
where $\hat{H}$ is the ground-truth edge heatmaps, ${\tt sg}(\cdot)$ stands for the stop-gradient operator, and $\beta$ is a hyper-parameter used for loss balance. The complete loss function for learning the codebook heatmap generator $\mathcal{L}_\text{codebook}$ is given by
\begin{equation}\label{equation:codebook total loss}
    \mathcal{L}_\text{codebook} = \mathcal{L}_\text{img} + \lambda_\text{latent} \cdot \mathcal{L}_\text{latent},
\end{equation}
where $\lambda_\text{latent}$ is a hyper-parameter used for loss balance.

\subsection{\ourmethod} \label{section:\ourmethod}

Given an occluded or partially non-visible face as input, conventional nearest-neighbor searching described in \eqref{equation:code replacement} may fail on the occluded patches due to their feature corruption.
However, relying solely on self-attention in transformers, \eg \cite{zhou2022towards}, is insufficient in heatmap generation since the attention map calculated with corrupted features no longer faithfully captures the relationships between patches.
To this end, we propose \ourmethod~to detect occluded patches and recover their features.

As shown in Figure~\ref{fig:overview}(b), we introduce the proposed \ourmethod~after training the heatmap generator.
\ourmethod~takes as input image patches $P=\{P_k\}_{k=0}^{m\times n-1}$ from the features $Z'$, which are extracted by the encoder $E$. 
\ourmethod~employs both regular and messenger tokens for computing patch features.
It generates the patch-specific occlusion map $\alpha \in \mathbb{R}^{m \times n}$ and two code sequences, $S_{I}\in \{0, 1, ..., N-1\}^{m \times n}$ and $S_{M}\in \{0, 1, ..., N-1\}^{m \times n}$.
While $S_{I}$ is computed from regular tokens and brings information from all patches, $S_{M}$ is derived from messenger tokens and is occlusion-aware.
Based on the codes in $S_{I}$ and $S_{M}$, quantized features $Z_{I}$ and $Z_{M}$ are produced by referring to codebook $C$. 
$Z_{I}$ and $Z_{M}$ are merged by based on the occlusion map $\alpha$ in a patch-specific manner, and form the recovered feature $Z_\text{rec}$. 
Finally, $Z_\text{rec}$ along with the pre-trained decoder is used to generate the heatmaps $H_\text{rec}$. 

We freeze the codebook $C$ and decoder $D$ after the pre-training stage while fine-tuning the encoder $E$ to facilitate heatmap generation under feature occlusion. 
The proposed \ourmethod, shown in Figure~\ref{fig:method}, is elaborated as follows:

\paragraph{Self-attention.}
As shown in Figure~\ref{fig:method}, \ourmethod~is a transformer with $L$ layers. 
At each layer $l$, it computes self-attention among regular image patch tokens by
\begin{equation}\label{equation:image token}
    X^{l+1} = {\tt FFN}\{{\tt softmax}(Q_X^l(K_X^l)^{\top})V_X^l + X^l\},
\end{equation}
where queries $Q_X^l$, keys $K_X^l$, and values $V_X^l$ are obtained from patch tokens $X^l$ through linear embeddings. 
Residual learning \cite{he2016deep} and a feed-forward network (FFN) are employed here.

\paragraph{Cross-attention.}
In addition to conventional self-attention between image patch tokens, we introduce the messenger tokens, denoted as $M^l$, one for each patch. 
The messenger tokens are designed to simulate feature occlusion. 
As shown in Figure~\ref{fig:method}, we only compute their queries $Q_M^l$, each of which is used to aggregate features from all but its corresponding patch token via cross-attention:
\begin{equation}\label{equation:messenger token}
    M^{l+1} = {\tt FFN}\{{\tt softmax}(A_\text{cross}(Q_M^l,K_X^l))V_X^l\},
\end{equation}
where
\begin{equation}\label{equation:cross-atten}
\small{
    A_\text{cross}(Q_M^l,K_X^l)^{i,j}=
        \begin{cases}
            0,& \text{if } i = j,\\
            (Q_M^l(K_X^l)^{\top})^{i,j},& \text{otherwise.}
        \end{cases}
    }
\end{equation}

\eqref{equation:cross-atten} computes the cross-attention score between the $i$-th \ourtoken~and the $j$-th image patch token. 
By excluding features from the corresponding patch, the resultant messenger tokens $M^{l+1}$ encode features borrowed from other image tokens, simulating feature occlusion.

\paragraph{Occlusion Detection Head.}
Following the attention mechanism and the feed-forward network, we derive an occlusion detection head to detect occluded patches by referring to the dissimilarity between the image patch embedding $X^{l+1}$ and the \ourembed~$M^{l+1}$. 
A patch-specific occlusion map $\alpha^{l+1} =\{\alpha^{l+1}_k\}_{k=0}^{m\times n-1}$ is obtained:
\begin{equation}
    \alpha^{l+1}_k = \sigma(W^{l+1} \cdot {\tt dist}(X^{l+1}_k, M^{l+1}_k)), 
\end{equation}
where the function ${\tt dist}(\cdot,\cdot)$ computes the element-wise squared difference between the two embeddings. 
$W^{l+1}$ is a fully connected layer transforming the embedding returned by ${\tt dist}$ into a scalar. 
$\sigma(*)$ is the sigmoid function ensuring $\alpha^{l+1}_k$ ranges between $0$ and $1$. 
Higher $\alpha^{l+1}_k$ indicates that patch $k$ is more likely to be occluded. 

\paragraph{Occlusion-aware Cross-attention.}
After obtaining the occlusion map $\alpha^l \in \mathbb{R}^{m \times n}$ at the $(l-1)$-th layer, the messenger tokens at the $l$-th layer are allowed to suppress feature aggregation from occluded patches.
Specifically, the cross-attention adopted by messenger tokens in \eqref{equation:cross-atten} is modified to
\begin{equation}\label{equation:cross-atten with alpha}
\small{
    A_\text{cross}(Q_M^l,K_X^l)^{i,j}=
        \begin{cases}
        0,& \text{if } i = j,\\
        (1-\alpha_j^l)(Q_M^l(K_X^l)^{\top})^{i,j},& \text{otherwise.}
        \end{cases}}
\end{equation}
Since $\alpha_j^{l}$ gives the likelihood of occlusion occurrence in the $j$-th image patch, the coefficient $(1-\alpha_j^{l})$ in \eqref{equation:cross-atten with alpha} prevents a messenger token from aggregating features from patch $j$ with a larger value of $\alpha_j^{l}$. 
At the first layer, the initial occlusion map $\alpha^{1}$ is set to $0$. 
At the last layer, \ie, the $L$-th layer, the resultant occlusion map $\alpha^{L+1}$ will be used in the following step for feature recovery, and is denoted as $\alpha$ for simplicity in Figure~\ref{fig:overview}(b).

\paragraph{Feature Recovery.}
In Figure~\ref{fig:method}, the image embedding $X^{L+1}$ and the \ourembed~$M^{L+1}$ produced in the last layer of \ourmethod~are fed into a codebook prediction head. 
This head predicts the code sequence $S_{I} \in \{0, 1, ..., N-1\}^{m \times n}$ based on the image embedding $X^{L+1}$, where each entry in $S_{I}$ searches the code index for its corresponding patch via \eqref{equation:code replacement}.
The quantized features $Z_I \in \mathbb{R}^{m \times n \times d}$ are produced by retrieving the corresponding $m \times n$ code items from the codebook $C$.
Similarly, the other code sequence $S_{M}$ and quantized features $Z_M$ are generated based on the \ourembed~$M^{L+1}$. 

\begin{figure}[t!]
    \centering
    \scalebox{0.52}{\includegraphics[width=0.9\textwidth]{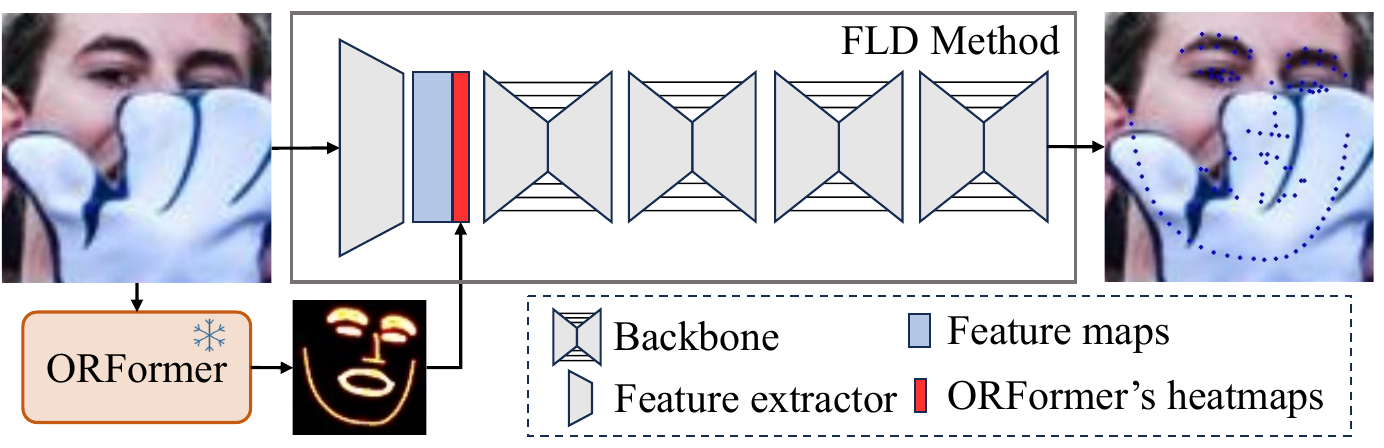}
    }
    \caption{
    \textbf{Integration of \ourmethod~into an existing FLD method.}
    \ourmethod~is adopted for occlusion detection and feature recovery, resulting in high-quality heatmaps. 
    The generated heatmaps serve as an extra input to an FLD method, and offer the recovered features to make the FLD method robust to occlusions.
    }
    \label{fig:integration}
\end{figure}

The quantized features $Z_I$ and $Z_M$ store complementary information.
While $Z_I$ considers all patches but is sensitive to corrupted features, $Z_M$ focuses on non-occluded patches but ignores the original patch features $P$ as shown in Figure~\ref{fig:method}.
We use the predicted occlusion map $\alpha \in \mathbb{R}^{m \times n}$ to recompose the final recovered features $Z_\text{rec} \in \mathbb{R}^{m \times n \times d}$ by merging $Z_I$ and $Z_M$ in a patch-specific manner, \ie,
\begin{equation}\label{equation:recomposition}
    Z_\text{rec} = (1-\alpha) \otimes Z_I + \alpha \otimes Z_M,
\end{equation}
where $A \otimes B$ denotes element-wise multiplication between $A$ and $B$ along the third dimension of $B$.

\paragraph{Loss Functions.}
 After the pre-training stage, we learn the \ourmethod~and fine-tune the encoder $E$ while keeping the codebook $C$ and the decoder $D$ fixed. 
We employ the cross-entropy loss for code sequence prediction $\mathcal{L}_\text{code}$ on both $S_I$ and $S_M$ via
\begin{equation}\label{equation:cross-entropy}
    \mathcal{L}_\text{code}(\hat{S}) = \sum_{k=0}^{m\times n-1}-S_klog(\hat{S_k}),
\end{equation}
where $\hat{S} \in \{S_I, S_M\}$ and the ground truth of the code sequence $S$ is obtained from the pre-trained heatmap generator mentioned in Section~\ref{section:VQVAE}. 
We also employ image-level loss $\mathcal{L}_\text{img}$ given in \eqref{equation:codebook loss} between $H_\text{rec}$ and $\hat{H}$. 
The complete loss function for learning  \ourmethod~ $\mathcal{L}_\text{\ourmethod}$ is
\begin{equation}\label{equation:\ourmethod total loss}
    \mathcal{L}_\text{\ourmethod} = \mathcal{L}_\text{code}(S_I) + \mathcal{L}_\text{code}(S_M) + \lambda_\text{img} \cdot \mathcal{L}_\text{img},
\end{equation}
where $\lambda_\text{img}$ is a hyper-parameter used for loss balance.

\subsection{Integration with FLD Methods} \label{section:FLD method}

With our \ourmethod~for occlusion detection and feature recovery, the quantized heatmap generator can produce high-quality heatmaps.
To evaluate the effectiveness of the output heatmaps, we integrate them as additional structural guidance into existing FLD methods \cite{huang2021adnet, zhou2023star}.
As illustrated in Figure~\ref{fig:integration}, the integration involves merging the heatmaps produced by the heatmap generator and the feature maps yielded by an existing FLD method.
Specifically, we concatenate the heatmaps with the feature maps in the early stage and merge them with a single lightweight CNN block.
Utilizing the merged features, our proposed method can model a more robust facial structure and enhance the performance of existing FLD methods, especially on occluded or partially non-visible faces.

\section{Experiments}\label{section:experiments}
\subsection{Experimental Settings}
\begin{table}[t!]
    \centering
    \scalebox{0.58}{
        \begin{tabular}{llccccccc}
            \toprule
            \multirow{2}{*}{Method}&\multirow{2}{*}{Backbone}&\multicolumn{3}{c}{WFLW-Full}&\multicolumn{1}{c}{COFW}&\multicolumn{3}{c}{300W (NME$\downarrow$)}\\
            \cmidrule(r){3-5}\cmidrule(r){6-6}\cmidrule(r){7-9}
            &&NME$\downarrow$&FR$\downarrow$&AUC$\uparrow$&NME$\downarrow$&Full&Comm.&Chal.\\
            \midrule
            LAB\cite{wu2018look}&Hourglass&5.27&7.56&0.532&-&3.49&2.98&5.19\\
            Wing\cite{feng2018wing}&ResNet-50&4.99&6.00&0.550&5.44&-&-&-\\
            HRNet\cite{sun2019deep}&HRNet-W18&4.60&4.64&-&-&3.32&2.87&5.15\\
            AWing\cite{wang2019adaptive}&Hourglass&4.36&2.84&0.572&4.94&3.07&2.72&4.52\\

            LUVLi\cite{kumar2020luvli}&DU-Net&4.37&3.12&0.577&-&3.23&2.76&5.16\\
            ADNet\cite{huang2021adnet}&Hourglass&4.14&2.72&0.602&4.68&\textcolor{blue}{2.93}&\textcolor{blue}{2.53}&4.58\\
            PIPNet\cite{jin2021pixel}&ResNet-101&4.31&-&-&-&3.19&2.78&4.89\\
            HIH\cite{lan2021hih}&Hourglass&4.08&2.60&0.605&4.63&3.09&2.65&4.89\\
            SLPT\cite{xia2022sparse}&HRNet-W18&4.14&2.76&0.595&4.79&3.17&2.75&4.90\\
            RePFormer\cite{li2022repformer}&ResNet-101&4.11&-&-&-&3.01&-&-\\
            \dag STAR\cite{zhou2023star}&Hourglass&4.03&\textcolor{blue}{2.32}&0.611&\textcolor{blue}{4.62}&\textcolor{red}{2.90}&\textcolor{red}{2.52}&\textcolor{blue}{4.46}\\
            LDEQ\cite{micaelli2023recurrence}&Hourglass&\textcolor{blue}{3.92}&2.48&\textcolor{red}{0.624}&-&-&-&-\\
            \midrule
            \ourmethod~(Ours)&Hourglass&\textcolor{red}{3.86}&\textcolor{red}{1.76}&\textcolor{blue}{0.622}&\textcolor{red}{4.46}&\textcolor{red}{2.90}&\textcolor{blue}{2.53}&\textcolor{red}{4.43}\\
            \bottomrule
        \end{tabular}
        }
    \caption{\textbf{Quantitative comparison with state-of-the-art methods on WFLW, COFW, and 300W.} NME is reported for all datasets. For WFLW, FR and AUC with a threshold of 10\% are included. The \textcolor{red}{best} and \textcolor{blue}{second best} results are highlighted. The \dag~symbol represents the results we reproduced.}
    \label{tab:table1}
\end{table}
\paragraph{Implementation Details.} 
The quantized heatmap generator takes images of resolution 64$\times$64$\times$3 as input and outputs 64$\times$64$\times$$N_E$ heatmaps, where $N_E$ is the number of edges per face.
Its latent space size is $m\times n\times d$, and the codebook size is $N\times d$, where $N$ is the number of code entries, and each entry is a $d$-dimensional vector.
The \ourmethod~is a 3-layer transformer structure operating within the latent space of the quantized heatmap generator.
Its token size is set to 1$\times$1$\times$$d$ with a total of $m\times n$ tokens.
We empirically set $m$ and $n$ to 16, $d$ to 256, and $N$ to 2,048.

For the landmark detection models, we follow the same setup as ADNet\cite{huang2021adnet} and STAR\cite{zhou2023star}. 
We use a four-stacked hourglass network \cite{newell2016stacked} as the backbone. 
Each hourglass outputs feature maps of resolution 64$\times$64. 
To incorporate the output heatmaps from \ourmethod~into the feature maps, we concatenate them and then apply a 1$\times$1 convolutional block to merge them before the first hourglass block. We train the model from scratch only on the target dataset without external data or pre-trained weights.
For the loss balance hyper-parameters, we empirically set $\beta$ to 0.25, $\lambda_\text{latent}$ to 100, and $\lambda_\text{img}$ to 50.

\begin{figure}[t!]
\centering
    \scalebox{0.45}{
    \includegraphics{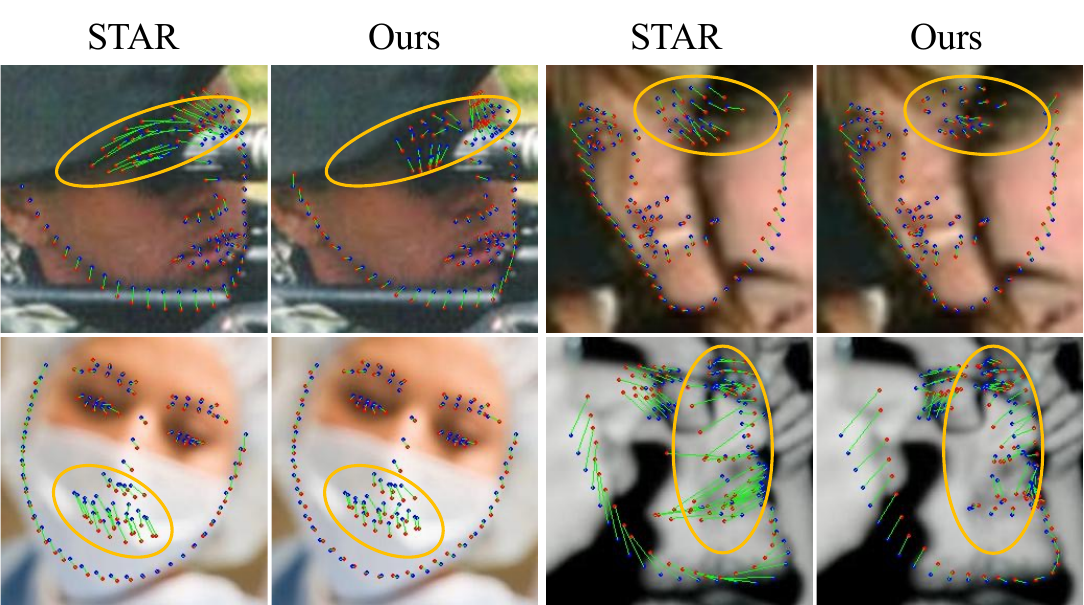}
    }
    \caption{\textbf{Qualitative comparison with the reproduced baseline method, STAR, on extreme cases from the test set of WFLW.} 
    The ground-truth landmarks are marked in blue, while the predicted landmarks are in red. The green lines represent the distance between the ground-truth landmarks and the predicted landmarks. 
    Orange ellipses highlight variations between the methods in the challenging areas.}
    \label{fig:FLD}
\end{figure}

\paragraph{Datasets.} 
We conducted experiments on three public datasets: WFLW \cite{wu2018look}, COFW \cite{burgos2013robust}, and 300W \cite{sagonas2013300}.

\textbf{WFLW} is widely recognized as the standard benchmark in facial landmark detection, containing 7,500 training images and 2,500 testing images, with 98 landmarks per image. 
This dataset presents significant challenges due to its diverse range of extreme cases, including variations in pose, expression, illumination, makeup, occlusion, and blur. 
Each sample in the dataset is accompanied by additional labels indicating the specific extreme case it represents.
       
\textbf{COFW} is a benchmark with 1,345 training images and 507 testing images, with 29 landmarks per image. 
This dataset is known for face occlusion, with an average of 23\% of landmarks occluded. 
Every landmark is accompanied by an additional label indicating whether the landmark is occluded or not.
However, we do not use the occlusion label in the experiments.

\textbf{300W} is commonly used in facial landmark detection, comprising 3148 training images and 689 testing images, with 68 landmarks per image. 
We also adopt the common setting on 300W, splitting the test set into the common subset of 554 images and the challenging subset of 135 images.

\paragraph{Data Augmentation.} 
We employ various data augmentation techniques in our experiments. For the quantized heatmap generator and \ourmethod, we start by cropping the face region from the original image and resizing it to 64$\times$64 pixels. Subsequently, we apply random horizontal flipping (50\%), random grayscale (50\%), random rotation ($\pm30^\circ$), random translation ($\pm4\%$), and random scaling ($\pm5\%$). Additionally, to enhance \ourmethod's ability to handle occlusions, we add an extra random occlusion to every image.

For the landmark detection model, we resize the images to 256$\times$256 pixels and set the random occlusion probability to 40\%, with other augmentation techniques kept.

\subsection{Evaluation Metrics}
Following previous works \cite{huang2021adnet, zhou2023star}, we employ three commonly used evaluation metrics to assess the accuracy of landmark detection: including normalized mean error (NME), failure rate (FR), and area under curve (AUC). 
For WFLW and 300W, inter-ocular NME is used, while for COFW, inter-pupil NME is used. 
For FR and AUC, the threshold is set to 10\% for all datasets.
       
\begin{table}[t!]
    \centering
    \scalebox{0.64}{
        \begin{tabular}{llc|cccccc}
            \toprule
            \multirow{2}{*}{Method}&\multirow{2}{*}{BackBone}&Full&Pose&Exp.&Ill.&Mak.&Occ.&Blur\\
            \cmidrule{3-9}
            &&\multicolumn{7}{c}{NME$\downarrow$}\\
            \midrule
                LAB\cite{wu2018look}&Hourglass&5.27&10.24&5.51&5.23&5.15&6.79&6.12\\
                Wing\cite{feng2018wing}&ResNet-50&4.99&8.43&5.21&4.88&5.26&6.21&5.81\\
                HRNet\cite{sun2019deep}&HRNet-W18&4.60&7.94&4.85&4.55&4.29&5.44&5.42\\
                AWing\cite{wang2019adaptive}&Hourglass&4.36&7.38&4.58&4.32&4.27&5.19&4.96\\
                ADNet\cite{huang2021adnet}&Hourglass&4.14&6.96&4.38&4.09&4.05&5.06&4.79\\
                PIPNet\cite{jin2021pixel}&ResNet-101&4.31&7.51&4.44&4.19&4.02&5.36&5.02\\
                HIH\cite{lan2021hih}&Hourglass&4.08&6.87&\textcolor{blue}{4.06}&4.34&3.85&4.85&4.66\\
                SLPT\cite{xia2022sparse}&HRNet-W18&4.14&6.96&4.45&4.05&4.00&5.06&4.79\\
                RePFormer\cite{li2022repformer}&ResNet-101&4.11&7.25&4.22&4.04&3.91&5.11&4.76\\
                \dag STAR\cite{zhou2023star}&Hourglass&4.03&\textcolor{blue}{6.78}&4.26&\textcolor{blue}{3.97}&3.85&4.82&\textcolor{blue}{4.59}\\
                LDEQ\cite{micaelli2023recurrence}&Hourglass&\textcolor{blue}{3.92}&6.86&\textcolor{red}{3.94}&4.17&\textcolor{blue}{3.75}&\textcolor{blue}{4.77}&\textcolor{blue}{4.59}\\
                \midrule
                ORFormer (Ours)&Hourglass&\textcolor{red}{3.86}&\textcolor{red}{6.63}&4.13&\textcolor{red}{3.83}&\textcolor{red}{3.61}&\textcolor{red}{4.57}&\textcolor{red}{4.50}\\
            \bottomrule
        \end{tabular}
    }
    \caption{
        \textbf{Quantitative comparison with state-of-the-art methods on WFLW and its six subsets.} NME is reported for all subsets. The \textcolor{red}{best} and \textcolor{blue}{second best} results are highlighted. The \dag~symbol represents the results we reproduced.}
    \label{tab:table2}
\end{table}

\begin{figure}[t!]
\vspace{-4mm}
    \centering
    \scalebox{0.47}{
    \includegraphics[width=\textwidth]{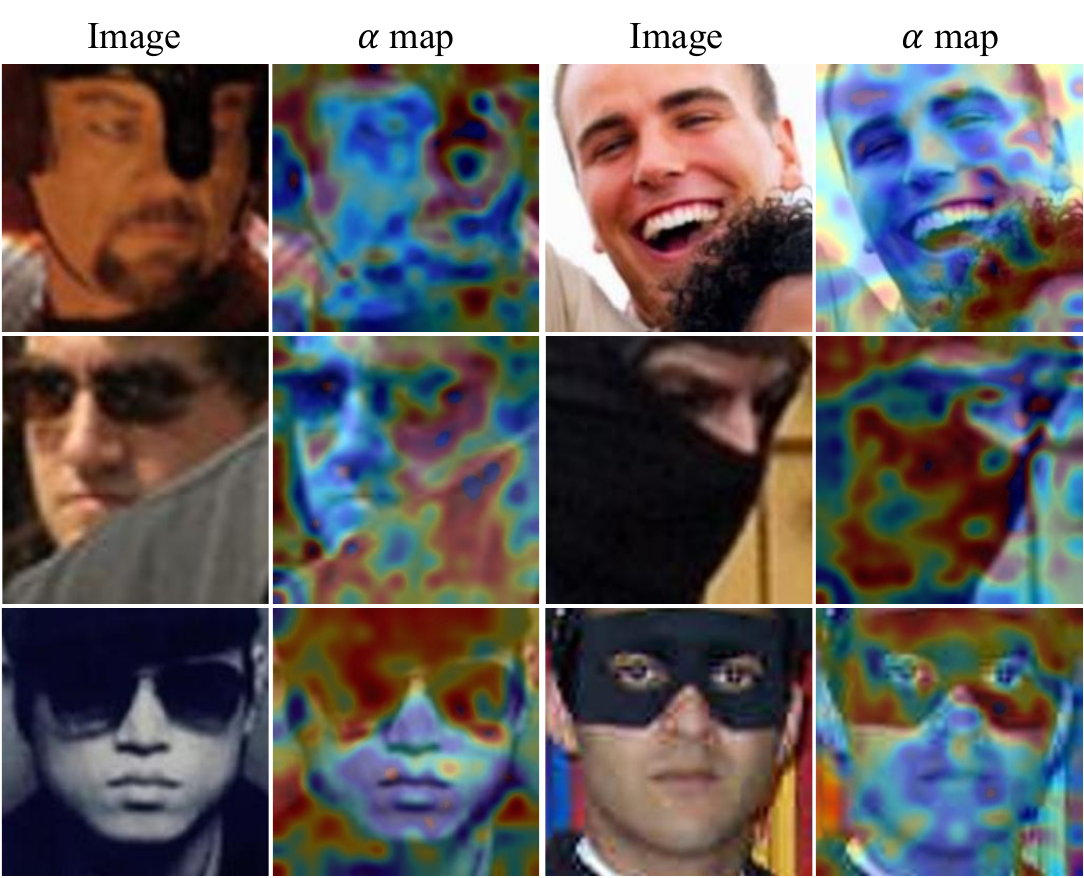}
    }
    \captionof{figure}{
    \textbf{Visualization of the $\alpha$ maps yielded by \ourmethod.} 
    Red regions indicate higher values of $\alpha$, suggesting heavier feature occlusion or corruption detected by \ourmethod.}
    \label{fig:Alpha}
\end{figure}
\subsection{Comparisons with State-of-the-Art Methods}
We conduct a comprehensive comparison of our method with state-of-the-art FLD approaches.
As presented in Table \ref{tab:table1}, \ourmethod~achieves comparable or even better results on the various facial landmark datasets.
Our method performs favorably against state-of-the-art methods in terms of NME and FR on WFLW, NME on COFW, and NME on the challenging subset of 300W, demonstrating the robustness of our \ourmethod~in challenging scenarios.
For qualitative comparison, we depict the output landmarks from our approach and a strong baseline method on randomly sampled images from the test set of WFLW.
We choose STAR \cite{zhou2023star} as our baseline method, which shares the same network architecture as ours.
We report the results reproduced using their official repository.
As shown in Figure~\ref{fig:FLD}, with the help of the high-quality heatmap generated from \ourmethod, our approach exhibits better robustness, particularly when the face is highly occluded or in extreme head rotations, demonstrating the capability of \ourmethod~to generate high-quality heatmaps resilient to extreme cases.

Furthermore, we analyze the performance across six subsets on the WFLW test set to validate the effectiveness of our method.
As presented in Table \ref{tab:table2}, \ourmethod~not only achieves state-of-the-art performance on the difficult subsets, but also excels in the occlusion, pose, illumination, and make-up subsets with large margins, demonstrating its robustness to partially non-visible facial features. The lower performance on the expression subset is attributed to the fact that it primarily consists of samples with deformed facial features.
\begin{figure}[t!]
\centering
    \scalebox{0.58}{
    \begin{tabular}{lc|cccccc}
        \toprule
        \multirow{2}{*}{Architecture}&Full&Pose&Exp.&Ill.&Mak.&Occ.&Blur\\
        \cmidrule{2-8}
        &\multicolumn{7}{c}{Heatmap L2 Loss $\downarrow$}\\
        \midrule
        VQVAE\cite{van2017neural}&26.72&51.43&29.32&26.82&30.96&35.63&32.07\\
        \multirow{2}{*}{CodeFormer\cite{zhou2022towards}}&25.01&49.61&27.98&25.22&27.30&32.46&30.08\\
        &(6.4\%)&(3.5\%)&(4.6\%)&(6.0\%)&\textbf{(11.8\%)}&\textbf{(8.9\%)}&(6.2\%)\\
        \midrule
        \multirow{2}{*}{\ourmethod~(Ours)}&20.22&36.51&23.80&20.83&21.65&25.94&24.70\\
        &(24.3\%)&\textbf{(29.0\%)}&(18.8\%)&(22.3\%)&\textbf{(30.1\%)}&\textbf{(27.2\%)}&(23.0\%)\\
        \bottomrule
    \end{tabular}
    }
\captionof{table}{\textbf{Quantitative comparison for heatmap generation on WFLW.} 
Heatmap regression L2 loss is reported for all subsets. 
The relative performance gain, given in parentheses, is calculated from the baseline VQVAE \cite{van2017neural}. 
Text in \textbf{bold} indicates a method gets a larger relative gain on that subset over the full set.}
    \label{tab:table3}
\end{figure}

Moreover, we visualize the output $\alpha$ map of \ourmethod~in Figure~\ref{fig:Alpha}. 
Regions in red indicate heavier occlusion detected by \ourmethod, while regions in blue denote less occlusion, highlighting our method's capability to detect occluded or corrupted features.

\begin{figure}[t!]
    \centering
    \scalebox{0.47}{
    \includegraphics[width=\textwidth]{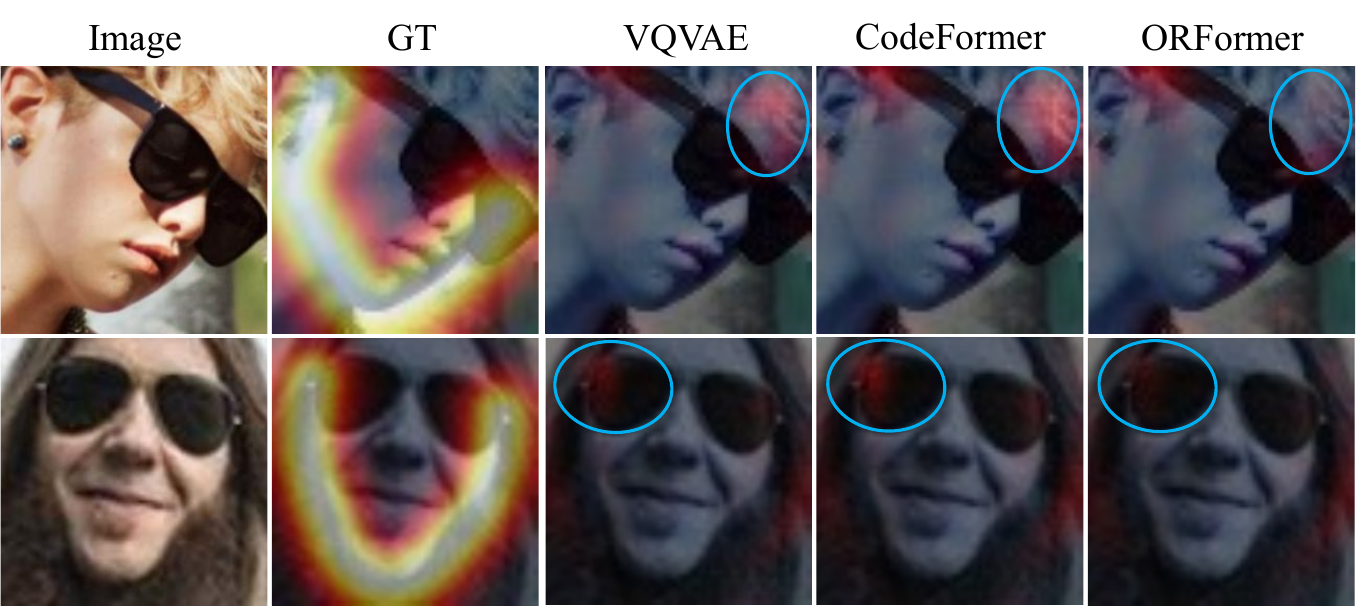}
    }
    \captionof{figure}{\textbf{Qualitative comparison for heatmap generation on WFLW.} GT stands for the ground-truth heatmap. 
    For better visualization, we display the distance heatmap for VQVAE, Codeformer, and \ourmethod~by computing the pixel-wise L2 distance between their output heatmaps and the GT heatmap. 
    Brighter areas indicate higher errors. 
    The main area of discrepancy is emphasized within an ellipse to highlight variations between the methods.}
    \label{fig:Heatmap}
\end{figure}

\subsection{Ablation Studies}
  
\paragraph{Effectiveness of \ourmethod.} 
To illustrate the performance difference of \ourmethod~in heatmap generation compared to baselines, we present the heatmap regression L2 loss on WFLW and its six subsets in Table \ref{tab:table3}.
We denote the quantized heatmap generator described in Section \ref{section:VQVAE} as VQVAE \cite{van2017neural} and use it as a baseline.
In addition, we denote the method that inserts a conventional ViT after the pre-trained encoder as CodeFormer \cite{zhou2022towards} for comparison.
While the method CodeFormer demonstrates performance gains over the baseline, \ourmethod~notably outperforms CodeFormer by a significant margin.
Particularly in the pose, make-up, and occlusion subsets, \ourmethod~exhibits exceptional performance compared to the full set, highlighting its robustness to feature occlusion and corruption.

We also visualize the output heatmap in Figure~\ref{fig:Heatmap}.
For ease of visual comparison, we choose the edge heatmap on the cheek and visualize the distance heatmap by computing the pixel-wise L2 distance between the output heatmap and the GT heatmap.
Brighter areas indicate larger errors.
In extreme cases, \ourmethod~can generate a better heatmap compared to other methods, showcasing the robustness to partial occlusion of \ourmethod.

\begin{table}[t]
    \centering
    \scalebox{0.54}{
    \begin{tabular}{l|c|c|c|c|c|c|c}
        \toprule
        Method&Param.&Self-Att.&Cross-Att.&Occ. Head&Occ.-Aware&L2 Loss $\downarrow$&NME$\downarrow$\\
        &(M)&&&&&&\\
        \midrule
        VQVAE \cite{van2017neural}&1.36&&&&&26.72&4.04\\
        CodeFormer \cite{zhou2022towards}&4.32&\checkmark&&&&25.13&4.00\\
        \midrule
        \multirow{3}{*}{ORFormer (Ours)}&4.77&\checkmark&\checkmark&&&24.35&3.99\\
        &4.78&\checkmark&\checkmark&\checkmark&&23.87&3.95\\
        &4.78&\checkmark&\checkmark&\checkmark&\checkmark&20.22&\textbf{3.86}\\
        \bottomrule
    \end{tabular}
    }
    \captionof{table}{\textbf{Quantitative evaluation on the proposed components of ORFormer on WFLW.} 
    The heatmap regression L2 loss and he landmark NME loss are reported.}
    \label{tab:table4}
\end{table}

\paragraph{Effectiveness of \ourmethod~Components.}
To assess the effectiveness of the components introduced in \ourmethod, we analyze the heatmap regression quality using heatmap L2 loss and the landmark detection accuracy using NME loss on WFLW. 
While VQVAE \cite{van2017neural} uses the quantized heatmap generator alone, CodeFormer \cite{zhou2022towards} integrates a traditional ViT within the latent space of the quantized heatmap generator. 
Table~\ref{tab:table4} reports the performance of each additional component based on the ViT \cite{dosovitskiy2020image} architecture.
We find that employing cross-attention with the proposed \ourtoken~provides an adequate improvement over CodeFormer \cite{zhou2022towards}.
With the occlusion detection head, our method equips the occlusion handling ability, reflecting on the drop of the L2 regression loss.
With our proposed occlusion-aware cross-attention, \ourmethod~effectively suppresses the feature aggregation from occluded patches, leading to a large margin in reducing the L2 loss and NME loss.

More ablation studies, the analysis of the computational complexity, and the discussion of the limitations of \ourmethod can be found in the supplementary materials.

\section{Conclusion}
In this paper, we introduce a novel occlusion-robust transformer architecture called \ourmethod, designed to detect occlusions and recover features for the occluded areas. 
Addressing the limitations of existing facial landmark detection (FLD) methods on difficult scenarios such as partially non-visible faces caused by occlusions, extreme lighting conditions, or extreme head rotations, \ourmethod~introduces new messenger tokens, which identify occluded features and recover missing details from visible observations. 
Through extensive ablation studies and experiments, we have demonstrated that \ourmethod~is able to generate high-quality heatmaps resilient to extreme cases. 
With the aid of \ourmethod, our method performs favorably against the state-of-the-art FLD methods on challenging datasets, such as WFLW and COFW.
\vspace{-3mm}
\paragraph{Ackknowledgements.} This work was supported in part by the National Science and Technology Council (NSTC) under grants 112-2221-E-A49-090-MY3, 111-2628-E-A49-025-MY3, 112-2634-F-006-002, and 113-2640-E-006-006. This work was funded in part by MediaTek.
\newpage
\section{Supplementary Materials}

We provide additional implementation details, more ablation studies, the analysis of the computational complexity, and the discussion of the limitations of \ourmethod~in this supplementary document.

\begin{figure*}[t]
\centering
\includegraphics[width=1\textwidth]{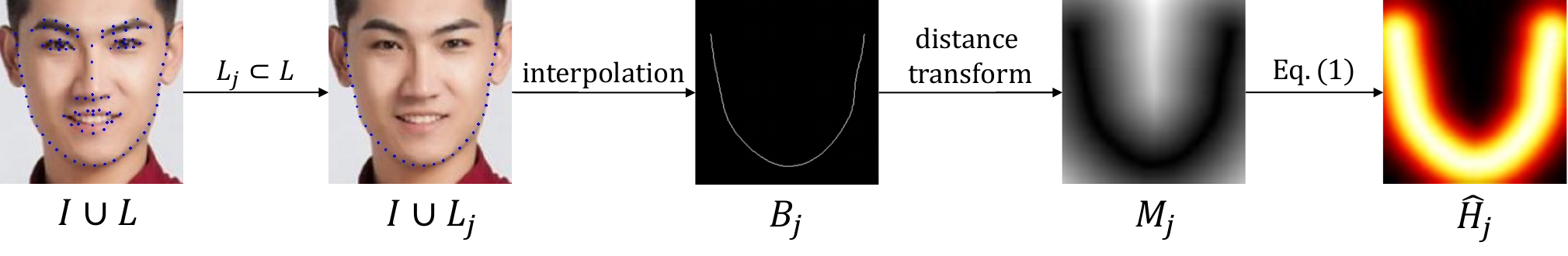}
\caption{\textbf{Generation flow of the ground-truth edge heatmap.}}
\label{fig:heatmap generation}
\end{figure*}
\begin{figure*}[t]
\centering
\scalebox{0.8}{
\includegraphics[width=1\textwidth]{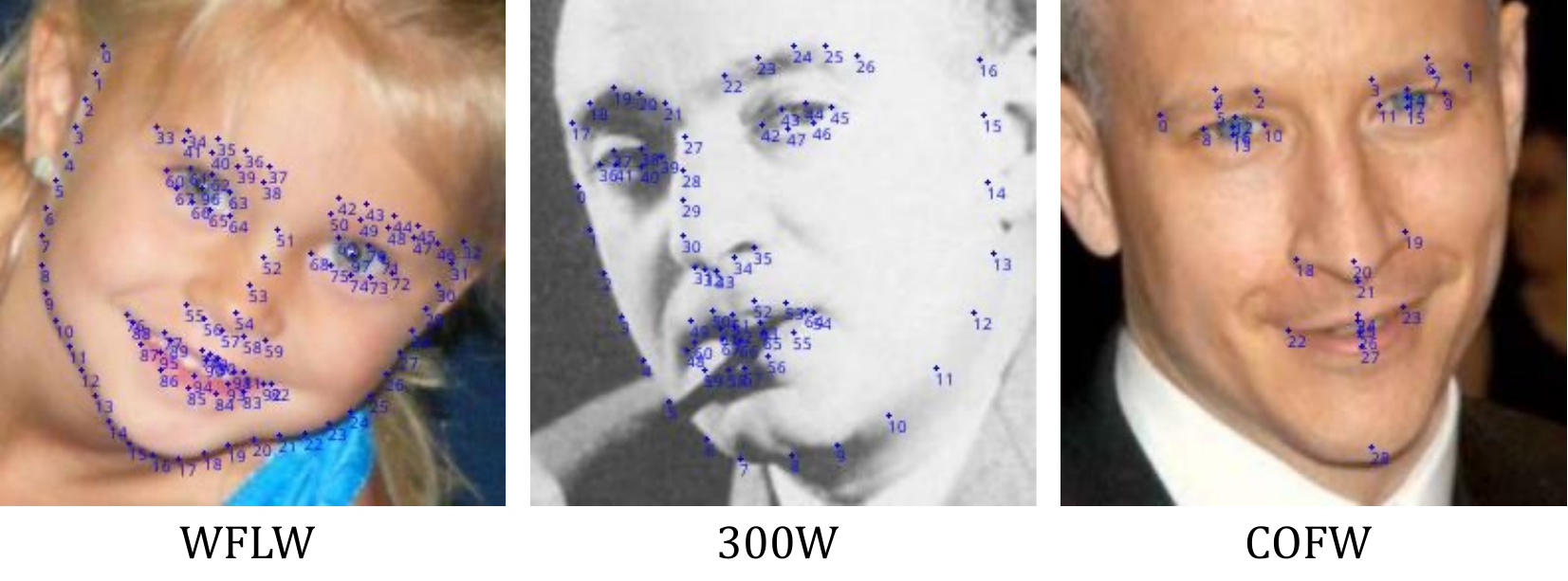}
}
\caption{\textbf{Visualization of the ground-truth landmarks in different datasets.}}
\label{fig:landmark}
\end{figure*}

\subsection{Additional Implementation Details}

\subsubsection{Model Training}
We employ the Adam optimizer \cite{kingma2014adam} along with the cosine annealing warm restart scheduler proposed by Loshchilov \etal \cite{loshchilov2016sgdr} in all our experiments. The number of iterations for the first restart is set to 5, and the increase factor is set to 2. 

The entire training process is carried out on a single NVIDIA GTX 1080 Ti with 11GB of memory. Specifically, for the quantized heatmap generator, we set the learning rate to 0.0007 with a batch size of 128. For deriving the proposed \ourmethod, we use a learning rate of 0.0001 with a batch size of 64. For the landmark detection models, we set the learning rate to 0.001 with a batch size of 16.

\subsubsection{Heatmap Definition}
\ourmethod~aims to identify non-visible regions and recover missing features, enabling the generation of high-fidelity heatmaps resilient to challenging scenarios like occlusions, extreme lighting conditions, or extreme head rotations. This capability assists facial landmark detection (FLD) methods in maintaining robustness in such challenging scenarios.

To support FLD methods effectively and efficiently, we employ heatmaps on facial edges (contours) as constraints by following a related approach proposed by Wu \etal \cite{wu2018look}. Utilizing edge heatmaps alone can reduce computational costs while providing sufficient information for FLD methods.

\paragraph{Heatmap Generation.}
As illustrated in Fig. \ref{fig:heatmap generation}, for a given face image $I \in \mathbb{R}^{h \times w \times 3}$ and its ground-truth landmark annotations $L=\{l_i\}_{i=0}^{N_L-1}$, we divide $L$ into $N_E$ subsets $L_j\subset L, j=0,...,N_E-1$ to represent the facial edges, such as the cheek and eyebrow. Here, $N_L$ represents the number of landmarks per face, and $N_E$ denotes the number of edges per face. Each facial edge $L_j$ is utilized to interpolate the edge line, thereby forming the binary edge map $B_j$ of the same size as the face image. Subsequently, a distance transform is applied to $B_j$, computing the nearest distance to the edge line for every pixel, resulting in the formation of the distance map $M_j$, which is also of the same size as the face image. Finally, we obtain the ground-truth edge heatmap $\hat{H_j}$ used to supervise the quantized heatmap and \ourmethod~by the following formula:
\begin{equation}
    \hat{H_j}(x,y) =
    \begin{cases}
        {\tt exp}(-\frac{M_j(x,y)^2}{2\sigma ^2}),& \text{if }M_j(x,y) < 3\sigma,\\
        0,& \text{otherwise.}
    \end{cases}
\end{equation}
$\sigma$ represents the standard deviation of the values in the distance map $M_j$.

\paragraph{Index Mapping.}

Our experiments are conducted on three distinct datasets: WFLW \cite{wu2018look}, COFW \cite{burgos2013robust}, and 300W \cite{sagonas2013300}. As illustrated in Fig. \ref{fig:landmark}, the number of landmarks varies across these datasets, leading to differences in the edge heatmap. Consequently, we provide the index mappings between the landmarks and the facial edges in the following.

\begin{table*}[t!]
    \centering
    \scalebox{0.8}{
        \begin{tabular}{llccccccc}
            \toprule
            \multirow{2}{*}{Method (Publication)}&\multirow{2}{*}{Backbone}&\multicolumn{3}{c}{WFLW-Full}&\multicolumn{1}{c}{COFW}&\multicolumn{3}{c}{300W (NME\textsubscript{io}$\downarrow$)}\\
            \cmidrule(r){3-5}\cmidrule(r){6-6}\cmidrule(r){7-9}
            &&NME\textsubscript{io}$\downarrow$&FR\textsubscript{10\%}$\downarrow$&AUC\textsubscript{10\%}$\uparrow$&NME\textsubscript{ip}$\downarrow$&Full&Comm.&Chal.\\
            \midrule
            LAB (CVPR18)\cite{wu2018look}&Hourglass&5.27&7.56&0.532&-&3.49&2.98&5.19\\
            Wing (CVPR18)\cite{feng2018wing}&ResNet-50&4.99&6.00&0.550&5.44&-&-&-\\
            HRNet (CVPR19)\cite{sun2019deep}&HRNet-W18&4.60&4.64&-&-&3.32&2.87&5.15\\
            AWing (ICCV19)\cite{wang2019adaptive}&Hourglass&4.36&2.84&0.572&4.94&3.07&2.72&4.52\\

            LUVLi (CVPR20)\cite{kumar2020luvli}&DU-Net&4.37&3.12&0.577&-&3.23&2.76&5.16\\
            ADNet (ICCV21)\cite{huang2021adnet}&Hourglass&4.14&2.72&0.602&4.68&\textcolor{blue}{2.93}&\textcolor{blue}{2.53}&4.58\\
            PIPNet (IJCV21)\cite{jin2021pixel}&ResNet-101&4.31&-&-&-&3.19&2.78&4.89\\
            HIH (arXiv21)\cite{lan2021hih}&Hourglass&4.08&2.60&0.605&4.63&3.09&2.65&4.89\\
            SLPT (CVPR22)\cite{xia2022sparse}&HRNet-W18&4.14&2.76&0.595&4.79&3.17&2.75&4.90\\
            RePFormer (arXiv22)\cite{li2022repformer}&ResNet-101&4.11&-&-&-&3.01&-&-\\
            \dag STAR (CVPR23)\cite{zhou2023star}&Hourglass&4.03&\textcolor{blue}{2.32}&0.611&\textcolor{blue}{4.62}&\textcolor{red}{2.90}&\textcolor{red}{2.52}&\textcolor{blue}{4.46}\\
            LDEQ (CVPR23)\cite{micaelli2023recurrence}&Hourglass&\textcolor{blue}{3.92}&2.48&\textcolor{red}{0.624}&-&-&-&-\\
            \midrule
            \ourmethod~(Ours)&Hourglass&\textcolor{red}{3.86}&\textcolor{red}{1.76}&\textcolor{blue}{0.622}&\textcolor{red}{4.46}&\textcolor{red}{2.90}&\textcolor{blue}{2.53}&\textcolor{red}{4.43}\\
            \bottomrule
        \end{tabular}
        }
    \caption{\textbf{Quantitative comparison with state-of-the-art methods on WFLW, COFW, and 300W.} NME is reported for all datasets. For WFLW, FR and AUC with a threshold of 10\% are included. The \textcolor{red}{best} and \textcolor{blue}{second best} results are highlighted. The \dag~symbol represents the results we reproduced.}
    \label{tab:sup_table1}
\end{table*}

For the \textbf{WFLW} dataset, with $N_L$ equal to 98 and $N_E$ equal to 15, the index mapping is given as follows:
\begin{verbatim}
Edge 0: [0-32]
Edge 1: [33-37]
Edge 2: [38-41,33]
Edge 3: [42-46]
Edge 4: [46-49,50]
Edge 5: [51-54]
Edge 6: [55-59]
Edge 7: [60-64]
Edge 8: [64-67,60]
Edge 9: [68-72]
Edge 10: [72-75,68]
Edge 11: [76-82]
Edge 12: [82-87,76]
Edge 13: [88-92]
Edge 14: [92-95,88]
\end{verbatim}

For the \textbf{300W} dataset, with $N_L$ equal to 68 and $N_E$ equal to 13, the index mapping is given as follows:
\begin{verbatim}
Edge 0: [0-16]
Edge 1: [17-21]
Edge 2: [22-26]
Edge 3: [27-30]
Edge 4: [31-35]
Edge 5: [36-39]
Edge 6: [39-41,36]
Edge 7: [42-45]
Edge 8: [45-47,42]
Edge 9: [48-54]
Edge 10: [54-59,48]
Edge 11: [60-64]
Edge 12: [64-67,60]
\end{verbatim}

For the \textbf{COFW} dataset, with $N_L$ equal to 29 and $N_E$ equal to 14, the index mapping is given as follows:
\begin{verbatim}
Edge 0: [0,4,2]
Edge 1: [2,5,0]
Edge 2: [1,6,3]
Edge 3: [3,7,1]
Edge 4: [8,12,10]
Edge 5: [10,13,8]
Edge 6: [9,14,11]
Edge 7: [11,15,9]
Edge 8: [18,21,19]
Edge 9: [20,21]
Edge 10: [22,26,23]
Edge 11: [23,27,22]
Edge 12: [22,24,23]
Edge 13: [23,25,22]
\end{verbatim}
\begin{figure*}[t!]
\centering
    \scalebox{0.6}{
    \includegraphics{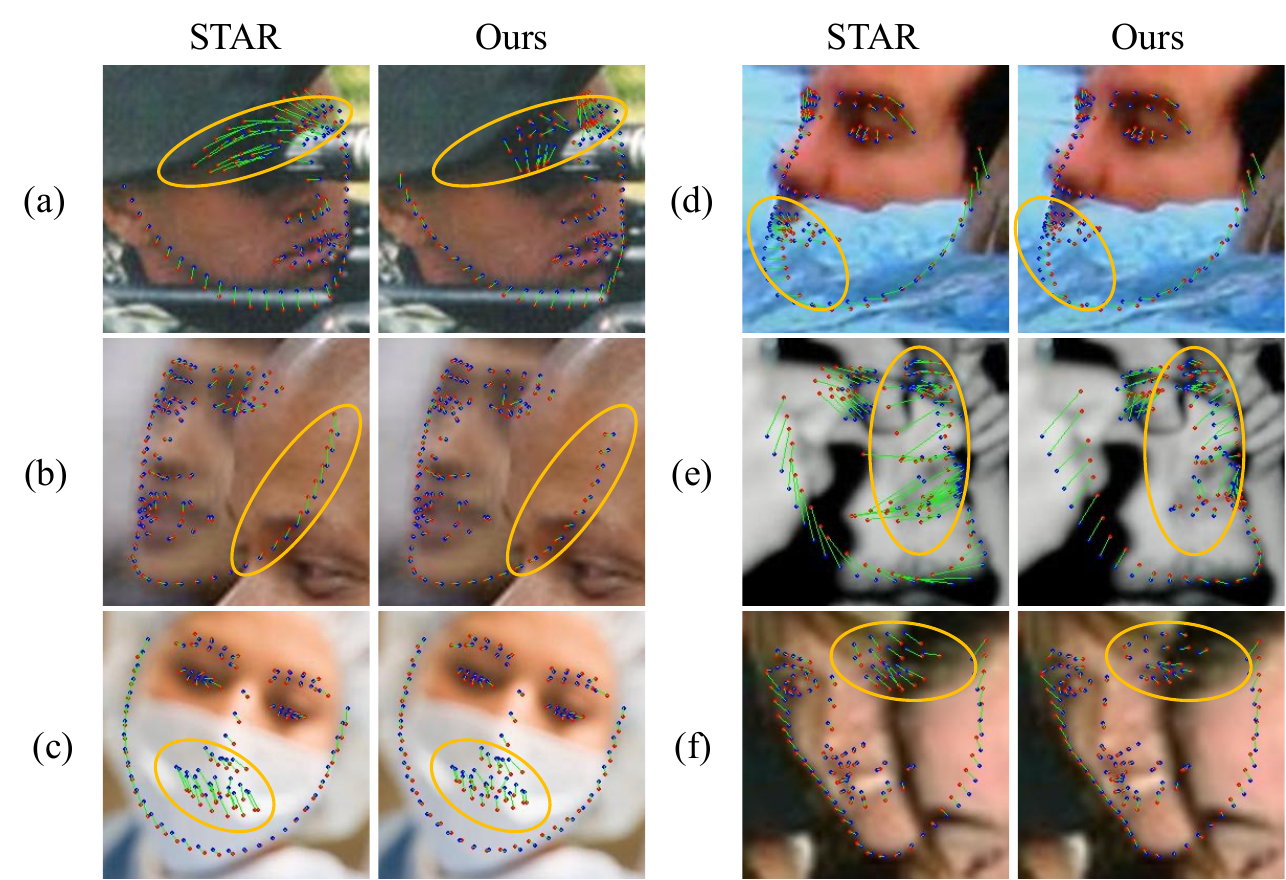}
    }
    \caption{\textbf{Qualitative comparison with the reproduced baseline method, STAR, on extreme cases from the test set of WFLW.} 
    The ground-truth landmarks are marked in blue, while the predicted landmarks are in red. The green lines represent the distance between the ground-truth landmarks and the predicted landmarks. 
    Orange ellipses highlight variations between the methods in the challenging areas.}
    \label{fig:sup_FLD}
\end{figure*}
\begin{figure*}[t!]
    \centering
    \scalebox{0.7}{
    \includegraphics[width=\textwidth]{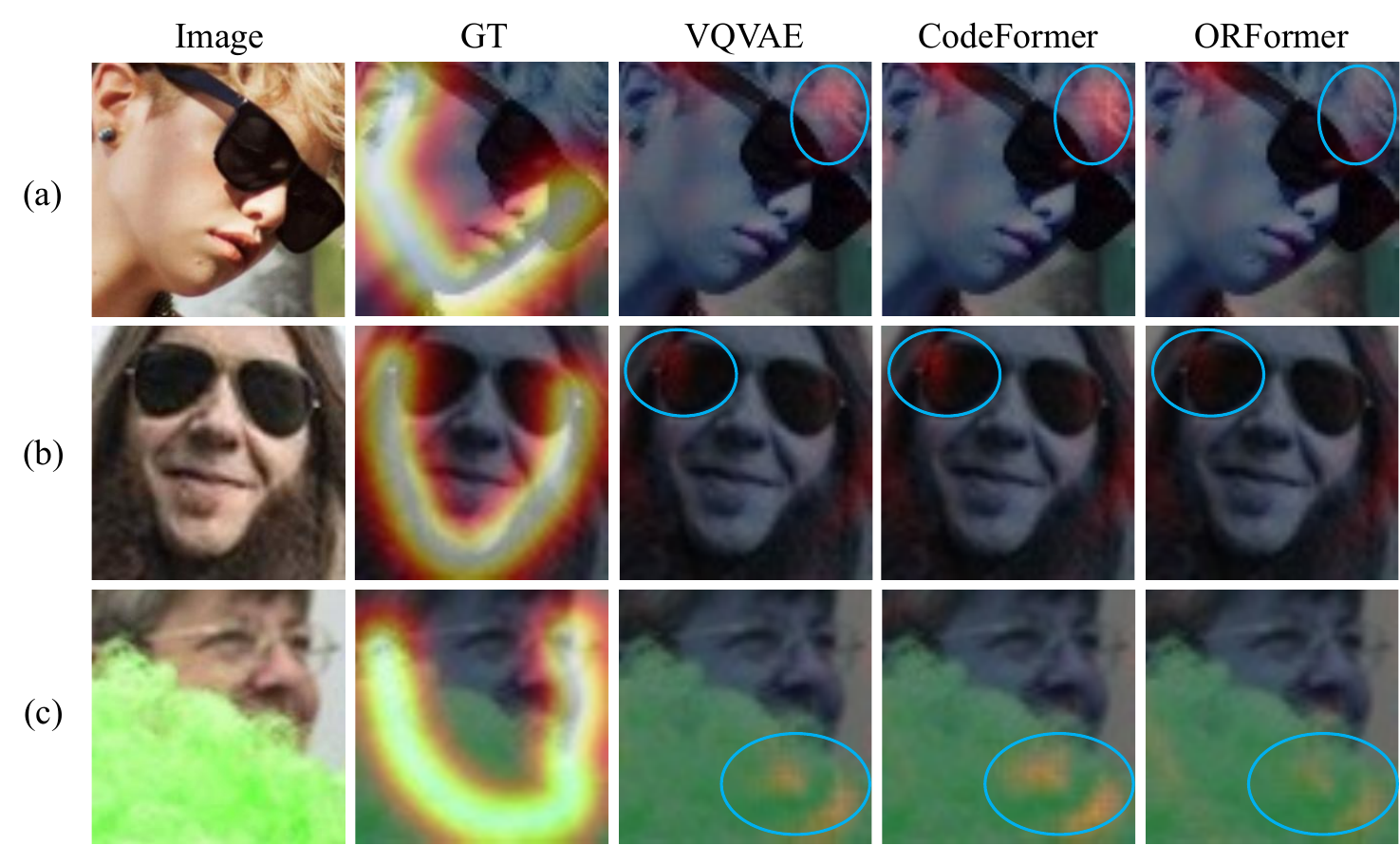}
    }
    \captionof{figure}{\textbf{Qualitative comparison for heatmap generation on WFLW.} GT stands for the ground-truth heatmap. 
    For better visualization, we display the distance heatmap for VQVAE, Codeformer, and \ourmethod~by computing the pixel-wise L2 distance between their output heatmaps and the GT heatmap. 
    Brighter areas indicate higher errors. 
    The main area of discrepancy is emphasized within an ellipse to highlight variations between the methods.}
    \label{fig:sup_Heatmap}
\end{figure*}

\subsection{More Experiments}
\subsubsection{Comparisons with State-of-the-Art Methods}
Due to limited space in the main paper, we provide the full experimental table here, as shown in Table \ref{tab:sup_table1}. We also provide more samples for visualization of the output landmark, as shown in Figure \ref{fig:sup_FLD}.
\subsubsection{Ablation Study}
\paragraph{Effectiveness of \ourmethod~.}
Due to limited space in the main paper, we provide more samples for visualization of the output heatmap of \ourmethod, as shown in Figure \ref{fig:sup_Heatmap}.
\begin{table}[t!]
    \centering
    \scalebox{0.67}{
    \begin{tabular}{llcccc}
        \toprule
        \multirow{2}{*}{Method}&\multirow{2}{*}{Architecture}&\multicolumn{2}{c}{Loss Functions}&\multicolumn{2}{c}{WFLW (NME$\downarrow$)}\\
        \cmidrule(r){3-4}\cmidrule(r){5-6}
        &&Heatmap&Landmark&Full&Occ.\\
        \midrule
        \multirow{2}{*}{ADNet\cite{huang2021adnet}}&HGNet&AWing&ADL&4.14&5.06\\
        &HGNet\texttt{+}\textbf{\ourmethod}&AWing&ADL&\textbf{4.06}&\textbf{4.94}\\
        &&&&(\texttt{+}1.9\%)&(\texttt{+}2.4\%)\\
        \midrule
        \multirow{2}{*}{\dag STAR\cite{zhou2023star}}&HGNet&AWing&STAR&4.03&4.82\\
        &HGNet\texttt{+}\textbf{\ourmethod}&AWing&STAR&\textbf{3.92}&\textbf{4.66}\\
        &&&&(\texttt{+}2.7\%)&(\texttt{+}3.3\%)\\
        \bottomrule
    \end{tabular}
    }
    \caption{\textbf{Ablation study of enabling~\ourmethod~for landmark detection on WFLW.} NME is reported. The \dag~symbol represents the results we reproduced. The relative performance improvement is calculated based on HGNet.}
    \label{tab:table5}
\end{table}
\paragraph{Effectiveness of \ourmethod's Heatmaps.}
To demonstrate the effectiveness of \ourmethod~for heatmaps generation for facial landmark detection, we compare it to the methods that utilize the same baseline network: ADNet \cite{huang2021adnet} and STAR \cite{zhou2023star}.
The results are presented in Table \ref{tab:table5}.
By incorporating \ourmethod's output heatmaps as additional information to the networks, alongside the same loss functions used by ADNet and STAR, our method achieves performance gains, especially in the occlusion subset, showing the effectiveness of \ourmethod's heatmap to existing FLD methods.
\paragraph{Occlusion Detection Head.}
As mentioned in the paper, we incorporate an occlusion detection head in our proposed \ourmethod~to identify occluded patches by evaluating the dissimilarity between the image patch embedding $X^{l+1}$ and the messenger embedding $M^{l+1}$. The patch-specific occlusion map $\alpha^{l+1} =\{\alpha^{l+1}_k\}_{k=0}^{m\times n-1}$ is obtained via
\begin{equation}
\label{eq:1}
\alpha^{l+1}_k = \sigma(W^{l+1} \cdot {\tt dist}(X^{l+1}_k, M^{l+1}_k)),
\end{equation}
where ${\tt dist}(\cdot,\cdot)$ calculates the element-wise squared difference between the two embeddings, $W^{l+1}$ represents a fully connected layer that transforms the embedding returned by ${\tt dist}$ into a scalar, and $\sigma(*)$ is the sigmoid function ensuring $\alpha^{l+1}_k$ ranges between $0$ and $1$. 

\begin{table}[t!]
    \centering

    \begin{tabular}{l|c|c|c}
        \toprule
        Method&Distance Function&L2 Loss $\downarrow$\\
        \midrule
        \ourmethod&$\text{concat}\{X-M, M-X\}$&24.86\\
        \ourmethod&$|X-M|$&24.43\\
        \ourmethod&$(X-M)^2$&\textbf{23.87}\\
        \bottomrule
    \end{tabular}
    \captionof{table}{\textbf{Quantitative evaluation on different designs of the distance function of \ourmethod's occlusion detection head.} $X$ represents the image patch embedding and $M$ represents the \ourembed. The label Occ. Head denotes the proposed occlusion detection head. The proposed occlusion detection head is not enabled in the first-row entry. The occlusion-aware cross-attention component is not enabled here. Results highlighted in \textbf{bold} represent the best performance. The heatmap regression L2 loss is reported on WFLW.}
    \label{tab:distance function}
\end{table}
\begin{table}[t!]
    \centering
    \begin{tabular}{l|c|c}
        \toprule
        Method&$W$'s Design&L2 Loss $\downarrow$\\
        \midrule
        \ourmethod&$5\times 5$ conv.&24.26\\
        \ourmethod&$3\times 3$ conv.&24.05\\
        \ourmethod&Fully connected layer&\textbf{23.87}\\
        \bottomrule
    \end{tabular}
    \caption{\textbf{Quantitative evaluation with different filter sizes of $W$ in \ourmethod's occlusion detection head}. The occlusion-aware cross-attention component is not enabled here. Results highlighted in \textbf{bold} represent the best performance. The heatmap regression L2 loss is reported on WFLW.}
    \label{tab:occlusion head}
\end{table}
To explore the difference between designs of distance functions, we compare the heatmap regression quality using L2 loss with various designs of the distance function, as shown in Table \ref{tab:distance function}.
We observe that employing the squared difference as the distance function in the occlusion detection head yields the best performance.
This improvement can be attributed to the squared difference function's capability to impose a larger penalty when there is a large disparity between the image patch embedding $X^{l+1}$ and the messenger embedding $M^{l+1}$, while still enabling the gradient to propagate continuously.

To explore incorporating more information into \ourmethod~during occlusion detection, we compare the heatmap regression quality using L2 loss with different filter sizes of $W$ in Eq. \ref{eq:1}, as shown in Table \ref{tab:occlusion head}.
For the convolutional layer, we reshape the embedding back to $\mathbb{R}^{m \times n \times d}$ before applying the convolution operation. 
In contrast, for the fully connected layer, we pass the embedding one by one, equivalent to applying a $1\times 1$ convolutional layer in the shape of $\mathbb{R}^{m \times n \times d}$. 
Using a larger filter size for the convolutional layer allows the occlusion detection head to consider more information from neighboring embeddings when detecting occlusion.
However, we observe that using a fully connected layer performs best.
We believe this is because \ourmethod~operates in the latent space of the quantized heatmap generator, considering one single embedding in this latent space can provide an appropriate receptive field for occlusion detection in human faces.

\begin{table}[t!]
    \centering
    \scalebox{0.73}{
    \begin{tabular}{l|c|c|c|c}
        \toprule
        Method&Integration&Pre-trained Weights&Trainable Part&NME\textsubscript{io} $\downarrow$\\
        \midrule
        \ourmethod&&&&4.03\\
        \ourmethod&\checkmark&\checkmark&Conv.&4.01\\
        \ourmethod&\checkmark&\checkmark&All&3.94\\
        \ourmethod&\checkmark&&All&\textbf{3.86}\\
        \bottomrule
    \end{tabular}
    }
    \captionof{table}{\textbf{Quantitative evaluation of different integration methods of \ourmethod~with Existing FLD Methods} The Conv. label indicates the $1\times 1$ CNN block used to merge the heatmap generated by \ourmethod~with the feature maps of existing FLD methods. The first row entry represent reproducing STAR \cite{zhou2023star} without the integration with \ourmethod. Results highlighted in \textbf{bold} represent the best performance. The landmark detection NME loss is reported on WFLW.}
    \label{tab:integration}
\end{table}
\begin{table}[t!]
    \centering
    \scalebox{0.70}{
    \begin{tabular}{llcccc}
        \toprule
        Method&Architecture&\multicolumn{2}{c}{Loss Functions}&\multicolumn{2}{c}{WFLW (NME$\downarrow$)}\\
        \cmidrule(r){3-4}\cmidrule(r){5-6}
        &&Heatmap&Landmark&Full&Occ.\\
        \midrule
        ADNet\cite{huang2021adnet}&HGNet&AWing&\textcolor{blue}{ADL}&4.14&5.06\\
        Ours&HGNet\texttt{+}\textcolor{blue}{\ourmethod}&AWing&ADL&4.06&4.94\\
        \midrule
        \dag STAR\cite{zhou2023star}&HGNet&AWing&STAR&4.03&4.82\\
        Ours&HGNet\texttt{+}\textcolor{blue}{\ourmethod}&AWing&STAR&3.92&4.66\\
        \midrule
        Ours&HGNet\texttt{+}\textcolor{blue}{\ourmethod}&L2&NME&\textbf{3.86}&\textbf{4.57}\\
        \bottomrule
    \end{tabular}
    }
    \caption{\textbf{Quantitative evaluation of different loss functions of the integration of \ourmethod~with Existing FLD Methods.} All methods utilize the same backbone. Loss functions highlighted in \textcolor{blue}{blue} represent the proposed approaches of that work. Results highlighted in bold represent the best performance. The landmark detection NME loss is reported on the WFLW dataset. The \dag~symbol represents the results we reproduced.}
    \label{tab:loss function}
\end{table}
\paragraph{Integration with FLD Methods.}
With our \ourmethod~for occlusion detection and feature recovery, the quantized heatmap generator can produce high-quality heatmaps. We integrate these heatmaps as additional structural guidance into existing FLD methods \cite{huang2021adnet, zhou2023star}. Specifically, we concatenate the heatmaps with the feature maps in the early stage and merge them with a single lightweight $1\times 1$ CNN block.

\subparagraph{Way of Integration.}

To explore the best strategy of integrating the heatmap produced by \ourmethod~into existing FLD methods, we compare the landmark detection accuracy using NME loss with different integration strategies. The results are shown in Table \ref{tab:integration}.
The pre-trained weights are from reproducing STAR \cite{zhou2023star} with an NME of 4.03.
We find that by only fine-tuning the lightweight CNN block, we gain little performance with the help of \ourmethod's heatmap. 
However, if we fine-tune the entire network or train the entire network from scratch without using pre-trained weights, we can achieve larger performance gains.

\subparagraph{Loss Function.}
We also explore alternative choices of the loss function for model integration. 
As shown in Table \ref{tab:loss function}, by integrating the output heatmaps of \ourmethod~into existing FLD methods \cite{huang2021adnet, zhou2023star} and using the same loss functions, our approach achieves improved performance. 
Moreover, we obtain the best result using a simple loss function such as L2 loss for heatmap supervision and NME loss for landmark supervision.
We believe this is because our heatmap definition differs from ADNet and STAR.
While our heatmap is suitable for L2 loss, their heatmap is defined for the use of their proposed specific loss functions.

\begin{table}[t]
    \centering
    \scalebox{0.54}{
    \begin{tabular}{l|c|c|c|c|c|c|c}
        \toprule
        Method&Param.&Self-Att.&Cross-Att.&Occ. Head&Occ.-Aware&L2 Loss $\downarrow$&NME$\downarrow$\\
        &(M)&&&&&&\\
        \midrule
        VQVAE \cite{van2017neural}&1.36&&&&&26.72&4.04\\
        CodeFormer \cite{zhou2022towards}&4.32&\checkmark&&&&25.13&4.00\\
        \midrule
        \multirow{3}{*}{ORFormer (Ours)}&4.77&\checkmark&\checkmark&&&24.35&3.99\\
        &4.78&\checkmark&\checkmark&\checkmark&&23.87&3.95\\
        &4.78&\checkmark&\checkmark&\checkmark&\checkmark&20.22&\textbf{3.86}\\
        \bottomrule
    \end{tabular}
    }
    \captionof{table}{\textbf{Quantitative evaluation on the proposed components of ORFormer on WFLW.} 
    The heatmap regression L2 loss and he landmark NME loss are reported.}
    \label{tab:sup_table4}
\end{table}
\subsubsection{Computational Complexity of \ourmethod}
In Table \ref{tab:sup_table4}, we show the numbers of trainable parameters of \ourmethod. 
Compared to the conventional ViT, ORFormer enhances ViT to handle occlusions with minimal overhead, with about 10\% more trainable parameters.

Even though ORFormer doubles the token count of a regular ViT, the patch token and messenger token compute attention scores separately, affecting the computational complexity linearly and thus minimally impacting the inference time.
In Table \ref{tab:table6}, we integrate our proposed ORFormer into the state-of-the-art baseline, STAR\cite{zhou2023star}, a 4-stack Hourglass network. 
For fair comparison, we augment the baseline network with one additional stack to align the number of trainable parameters. 
Our method performs favorably against this augmented baseline with comparable trainable parameters, 20.6\% fewer mult-add operations, and 15.9\% less inference time, showing the advantage of \ourmethod.
\subsubsection{Limitations}
The first limitation is that \ourmethod~is particularly effective at handling partially non-visible facial features but struggles with partially deformed facial features.
The second limitation is that although \ourmethod~yields features robust to occlusion, the capability of our method relies on a well-trained quantized heatmap generator, which limits its applicability to tasks related to heatmap generation.
In future research, we plan to explore ways to enable \ourmethod~to handle partially deformed facial features and extend \ourmethod~to serve as a general feature extractor for various computer vision tasks, where partial occlusions detection and feature recovery are essential, maximizing its impact in the field of computer vision.
\begin{table}[t]
    \centering
    \scalebox{0.6}{
        \begin{tabular}{llccccc}
        \toprule
        \multirow{2}{*}{Method}&\multirow{2}{*}{Architecture}&Param.&Mult-Add&Infer. Time& NME\textsubscript{io} $\downarrow$\\
        &&(M)&(G)&(ms)&\\
        \midrule
        \dag STAR\cite{zhou2023star}&4-stack HGNet&17&17.4&45&4.03\\
        \dag STAR\cite{zhou2023star}&5-stack HGNet&21.5&21.5&63&3.98\\
        Ours&4-stack HGNet\texttt{+}ORFormer&21.8&17.9&53&\textbf{3.86}\\
        &&(\texttt{+}1.4\%)&(\texttt{-}20.6\%)&(\texttt{-}15.9\%)&\\
        \bottomrule
    \end{tabular}
    }
    \caption{\textbf{Ablation study of computation complexity vs NME on WFLW.} HGNet represents the hourglass network. The relative increase/improvement is calculated based on 5-stack HGNet. The \dag~symbol represents the results we reproduced. The inference time is tested on a single NVIDIA GTX 1080 Ti.}
    \label{tab:table6}
\end{table}

{\small
\bibliographystyle{ieee_fullname}
\bibliography{main}
}

\end{document}